\documentclass[10pt,twocolumn,letterpaper]{article}

\usepackage{cvpr}              % To produce the CAMERA-READY version
\definecolor{cvprblue}{rgb}{0.21,0.49,0.74}
\usepackage[pagebackref,breaklinks,colorlinks,allcolors=cvprblue]{hyperref}

\def\paperID{60} % *** Enter the Paper ID here
\def\confName{CVPR}
\def\confYear{2025}

\title{You Are Your Best Teacher: Semi-Supervised Surgical Point Tracking with Cycle-Consistent Self-Distillation}

\author{  Valay Bundele\quad
  Mehran Hosseinzadeh\quad
  Hendrik Lensch\\
  University of Tübingen\\
  {\tt\small \{valay.bundele, mehran.hosseinzadeh, hendrik.lensch\}@uni-tuebingen.de}
}

\begin{document}
\maketitle
\begin{abstract}
%Synthetic data has enabled significant progress in point tracking by providing large-scale, densely labeled training data, but adapting synthetic-trained models to real-world domains remains challenging due to domain shift and scene complexity. While recent methods adapt synthetic-pretrained models to real-world videos using pseudo-labels, their effectiveness in domains with significant distributional shift—such as surgical video—remains unexplored. Surgical video analysis exemplifies this difficulty: accurate point tracking is critical for modeling tissue dynamics and enabling automation, yet annotated data is unavailable, and factors such as tissue deformation, occlusions, and lighting variations further hinder performance. 
Synthetic datasets have enabled significant progress in point tracking by providing large-scale, densely annotated supervision. However, deploying these models in real-world domains remains challenging due to domain shift and lack of labeled data—issues that are especially severe in surgical videos, where scenes exhibit complex tissue deformation, occlusion, and lighting variation. While recent approaches adapt synthetic-trained trackers to natural videos using teacher ensembles or augmentation-heavy pseudo-labeling pipelines, their effectiveness in high-shift domains like surgery remains unexplored. This work presents SurgTracker, a semi-supervised framework for adapting synthetic-trained point trackers to surgical video using filtered self-distillation. Pseudo-labels are generated online by a fixed teacher—identical in architecture and initialization to the student—and are filtered using a cycle consistency constraint to discard temporally inconsistent trajectories. This simple yet effective design enforces geometric consistency and provides stable supervision throughout training, without the computational overhead of maintaining multiple teachers. Experiments on the STIR benchmark show that SurgTracker improves tracking performance using only 80 unlabeled videos, demonstrating its potential for robust adaptation in high-shift, data-scarce domains.
\end{abstract}    
\section{Introduction}
\label{sec:intro}
Tracking visual points over time is a core problem in computer vision, underpinning applications in motion understanding, visual correspondence, and robotic perception. Recent advances in learning-based point trackers~\cite{karaev2024cotracker, doersch2023tapir, li2024taptr, cho2024local} have shown remarkable performance by training on large-scale synthetic datasets with dense supervision. These models benefit from scalability and control in simulation, but transferring them to real-world scenarios remains a major challenge due to domain shift and lack of annotated data.

To mitigate this gap, recent efforts~\cite{doersch2024bootstap, karaev2024cotracker3} propose semi-supervised adaptation strategies using pseudo-labels generated on unlabeled natural videos. These methods leverage teacher-student frameworks and consistency losses to refine models in the absence of ground truth. However, they have been validated primarily on natural video domains, which, despite being unlabeled, still resemble the synthetic training distribution in terms of motion regularity and scene composition. Their applicability to more specialized, high-variance domains remains largely unexplored.

One such domain is surgical video analysis, where accurate point tracking can facilitate understanding of tissue dynamics, tool-tissue interaction, and intraoperative state estimation—critical for applications such as surgical skill assessment, automation, and guidance~\cite{schmidt2024tracking}. However, the domain poses unique challenges: deformable anatomy, specular lighting, heavy occlusion, and rapid motion. Moreover, obtaining annotated datasets for point tracking in surgery is impractical due to privacy concerns, the need for domain expertise, and the high cost of manual labeling. 

Prior methods in point tracking in surgical videos have typically relied on classical techniques such as sparse feature matching or optical flow~\cite{ihler2020patient}. Recent work such as SurgMotion~\cite{zhan2024tracking} adapts OmniMotion \cite{wang2023tracking} to surgical data using domain-specific priors, but requires test-time optimization, making it less practical for real-time deployment. As a result, the question remains: \textit{can recent synthetic-trained point trackers be effectively adapted to surgical video—without any manual annotations?}

To address this, we propose SurgTracker, a semi-supervised framework for adapting synthetic-trained point trackers to surgical video using only unlabeled data. While CoTracker3~\cite{karaev2024cotracker3} adapts to natural videos using pseudo-labels from diverse teacher models, we find that this approach is less effective in surgical settings, where the domain shift is more pronounced. Instead, SurgTracker employs a simpler yet more effective strategy: it leverages pseudo-labels from a single frozen teacher, identical to the student in architecture and initialization, and applies a cycle consistency constraint to retain only temporally coherent trajectories.

We attribute effectiveness of this design to three factors: first, diverse teachers introduce higher supervision variance due to inconsistent behaviors under domain shift, making pseudo-label quality less reliable; second, architectural alignment between teacher and student improves representational compatibility, allowing for more effective learning; and third, using a fixed teacher yields a stable supervisory signal across training batches, reducing fluctuations in optimization dynamics. In addition, our single-teacher setup eliminates the need to keep multiple large models in memory during training, making the approach more computationally efficient. Experiments on STIR benchmark~\cite{zhan2024tracking} show that SurgTracker improves tracking performance using only 80 unlabeled videos, demonstrating that in high-shift data-scarce domains, supervision consistency and alignment can outweigh benefits of teacher diversity.

\section{Related Works}
\subsection{Point Tracking}
% Recent point trackers such as PIPs, TAPIR, and CoTracker3 have achieved strong results by training on synthetic datasets with dense supervision. While synthetic data enables scale, it introduces a domain gap that affects generalization to real-world videos. To address this, some methods (e.g., BootsTAPIR) fine-tune on massive unlabeled video corpora using teacher-student self-training and heavy augmentations. More recent work like CoTracker3 demonstrates that adaptation with pseudo-labels from an ensemble of synthetic-trained teachers can yield similar gains with far less data. These approaches, however, have been mostly validated on natural video datasets and do not address more complex domains like surgical video.
Deep learning-based point trackers have advanced rapidly, largely by training on synthetic datasets due to the difficulty of labeling real-world trajectories. Early work like PIPs~\cite{harley2022particle} framed dense tracking as long-range motion estimation, later extended to longer sequences in PIPs++ \cite{zheng2023pointodyssey}. TAPIR \cite{doersch2023tapir} built on this by introducing global matching, while CoTracker~\cite{karaev2024cotracker} leveraged transformers to jointly track multiple points and better handle occlusion. More recent variants like LocoTrack~\cite{cho2024local} uses 4D correlation volumes whereas Track-On \cite{aydemir2025track} enables frame-by-frame tracking using spatial and context memory.
% Deep learning-based point trackers have advanced rapidly, often by training on synthetic video datasets due to the difficulty of labeling real-world point trajectories. Early work like PIPs \cite{harley2022particle} pioneered treating dense point tracking as a long-range motion estimation problem, enabling tracking of every point within a short time window. PIPs was later extended (PIPs++) to longer sequences \cite{zheng2023pointodyssey}, and the Tracking Any Point (TAP) benchmark (TAP-Vid) \cite{doersch2022tap} was introduced with annotated real video datasets. Building on these, TAPIR \cite{doersch2023tapir} improved PIPs by adding a global matching stage for more robust correspondence across frames. Another line, CoTracker \cite{karaev2024cotracker}, uses a transformer to jointly track multiple points and exploits cross-point correlations to better handle occlusions. Recent variants like LocoTrack \cite{cho2024local} further modify the architecture using 4D correlation volumes. 
While these methods show strong performance, they are trained on synthetic datasets and have been validated primarily on natural video domains.

%In contrast, the application of point tracking to surgical video remains underexplored. 
Point tracking in surgical videos is essential for modeling tissue dynamics and enabling image-guided robotic interventions~\cite{zhan2024tracking}. Classical methods based on sparse features or dense optical flow \cite{ihler2020patient} are limited by poor texture, deformation, and occlusion in surgical scenes. Recent approaches such as SENDD \cite{schmidt2023sendd} use graph-based models to jointly estimate 2D correspondences and 3D deformation. More recently, Zhan et al. \cite{zhan2024tracking} introduced a benchmark with manually annotated trajectories and proposed SurgMotion, which adapts OmniMotion \cite{wang2023tracking} with domain-specific priors. While effective, SurgMotion relies on test-time optimization, limiting its applicability in real-time settings. In contrast, our work explores whether synthetic-trained trackers can be adapted to surgical videos without any labels to enable robust real-time performance in clinical scenarios.

% Accurate tracking of tissue and instrument points in endoscopic videos is crucial for image-guided surgery and robotic assistance \cite{zhan2024tracking}. Traditional surgical tracking methods \cite{schmidt2024tracking} relied on either sparse feature correspondences or optical flow, but these have clear limitations. On one hand, feature-based methods struggle on deformable, texture-poor tissues and only provide sparse points with limited coverage. On the other hand, dense optical flow can capture finer motion but only between consecutive frames, making it prone to drift and failure under occlusions in long sequences \cite{ihler2020patient}. Recent specialized approaches have attempted to overcome these issues. For example, SENDD \cite{schmidt2023sendd} proposed graph-based models that match tissue keypoints and jointly estimate per-point depth and 3D deformation. Only very recently have general “track-any-point” methods been evaluated on surgical data. Zhan et al. (2024) introduced a surgical point tracking benchmark with manually annotated point trajectories and proposed SurgMotion which builds upon OmniMotion, by adding domain-specific losses. However, they introduce a test-time optimization technique which can't be used in real time.

\subsection{Unsupervised Domain Adaptation}
While synthetic data enables scalable training, domain shift remains a core challenge when deploying models on real-world videos. Self-training with pseudo-labels has emerged as a promising strategy, wherein source-trained models generate labels on unlabeled target data to guide fine-tuning. BootsTAP~\cite{doersch2024bootstap} applies this paradigm to large-scale natural video via teacher-student learning and strong augmentations. CoTracker3~\cite{karaev2024cotracker3} improves efficiency by distilling pseudo-labels from multiple teacher models,
%Self-training with pseudo-labels has shown to be effective in prior works, where source-trained models generate labels on target data for fine-tuning. BootsTAP ~\cite{doersch2024bootstap} applies this approach to large-scale natural videos using teacher-student learning with strong augmentations. CoTracker3 ~\cite{karaev2024cotracker3} improves efficiency by leveraging pseudo-labels from multiple teacher models 
but applies no filtering to account for label noise. Sun et al.~\cite{sun2024refining} incorporate cycle consistency to improve label quality, but compute pseudo-labels only once and keep them fixed, increasing susceptibility to confirmation bias.

Critically, these approaches have been validated only on natural videos, and it remains unclear whether they generalize to domains with significantly higher distribution shift—like surgical video. We address this gap by extending self-training-based point tracking to surgical data, leveraging a single, architecture-aligned teacher and applying cycle consistency filtering to provide  stable supervision.

\section{Method}
\begin{figure*}[t]  
    \centering \includegraphics[width=0.74\textwidth]{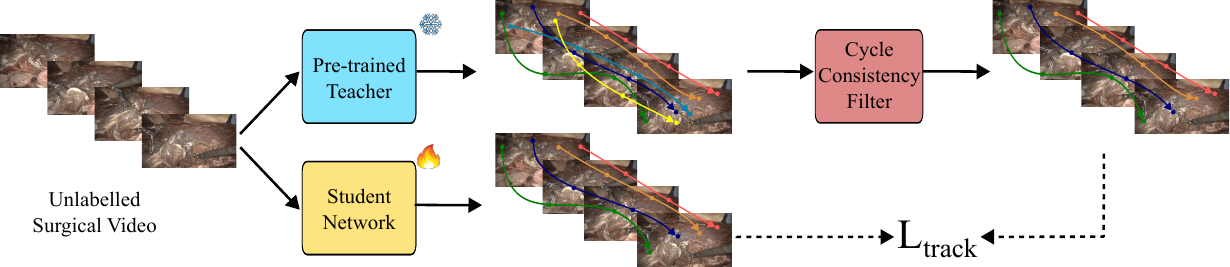}  
    \caption{Overview of the SurgTracker framework. Given an unlabeled surgical video, pseudo-labels are generated by a frozen teacher network and filtered using a cycle consistency check to remove temporally inconsistent trajectories. The filtered trajectories supervise the student model, which is fine-tuned using a tracking loss $\mathcal{L}_{\text{track}}$. The teacher model remains frozen during training. } 
    \label{fig:pipeline} 
\end{figure*}
\subsection{Problem Formulation}
Tracking tissue motion in surgical videos involves accurately following specific tissue points across frames. Given a video sequence $V = \{I_t\}_{t=1}^{T}$ consisting of $T$ frames, our objective is to track a set of $N$ query points $Q = \{(x_i, y_i, t_0)\}_{i=1}^{N}$ where $(x_i, y_i)$ denotes spatial location of $i$-th query point in the frame $t_0$. The goal is to estimate a trajectory $P = \{(x_i^t, y_i^t)\}_{t=1}^{T}$ for each query point $i$, representing its predicted location in every frame of the sequence.
\subsection{SurgTracker}
We propose SurgTracker, a semi-supervised framework for adapting synthetic-pretrained point trackers to surgical video, where large domain shift and lack of annotations present significant challenges. Our method leverages CoTracker3—pretrained on synthetic data and adapted to natural videos—as a fixed teacher to produce pseudo-labels, which are then filtered via a cycle consistency constraint to remove noisy trajectories. The student model, identical in architecture and initialization to the teacher, is then fine-tuned using these filtered labels. An overview of the SurgTracker pipeline is shown in Fig.~\ref{fig:pipeline}.

Unlike prior work that relies on teacher model ensembles \cite{karaev2024cotracker3} or large-scale data augmentation \cite{doersch2024bootstap}, SurgTracker uses a single teacher—architecturally aligned with the student—and leverages temporal consistency to identify high-quality training signals. This simple yet effective design enables adaptation to surgical videos without requiring any annotations. The method consists of three main stages: (1) pseudo-label generation, (2) trajectory filtering via cycle consistency, and (3) supervised fine-tuning of the student.
%, as pseudo-label generators. These models provide complementary strengths in handling occlusions, deformations, and motion consistency. The diversity in teacher predictions helps mitigate domain adaptation issues when transferring from natural to surgical video data. 

\subsubsection{Pseudo-Label Generation}
%We begin by sampling a set of query points $Q$ in the first frame of each training sequence. To ensure these points are informative and trackable, we use keypoints extracted using SIFT. 
For each training sequence, we sample a set of query points $Q$ from the first frame. To ensure that these points are informative and trackable, we extract keypoints using SIFT \cite{lowe1999object}, which provides robust features under appearance changes and viewpoint variation. Sequences with an insufficient number of detected keypoints are excluded to maintain supervision quality. The teacher model $M$ then predicts candidate trajectories $\hat{P}$ for each query point $q_i \in Q$. %Candidate trajectories $\hat{P}$ for the query points $Q$ are then predicted by the teacher model $M$. 

\subsubsection{Cycle-Consistent Filtering}
To improve the quality of pseudo-labels, we apply a cycle consistency check to identify and discard noisy trajectories. Let $\hat{P}_i = \{(x_i^t, y_i^t)\}_{t=t_0}^{t_1}$ denote the forward trajectory for a query point $q_i = (x_i^{t_0}, y_i^{t_0})$ generated by the teacher $M$, where $t_0$ and $t_1$ are the start and end frames of the sequence. We then perform reverse tracking starting from the final predicted location $(x_i^{t_1}, y_i^{t_1})$, obtaining a backward trajectory $\tilde{P}_i = \{(\tilde{x}_i^t, \tilde{y}_i^t)\}_{t=t_1}^{t_0}$, again using $M$. We define cycle consistency error as the Euclidean distance between original query point and endpoint of the backward track:

\begin{equation}
    \mathcal{E}_{\text{cycle}}(q_i) = \left\| (x_i^{t_0}, y_i^{t_0}) - (\tilde{x}_i^{t_0}, \tilde{y}_i^{t_0}) \right\|_2.
\end{equation}

A trajectory $\hat{P}_i$ is considered valid if the cycle consistency error satisfies $\mathcal{E}_{\text{cycle}}(q_i) < \alpha$, where $\alpha$ is a hyperparameter controlling the filtering aggressiveness. Only valid trajectories are used as pseudo-labels to supervise the student. Throughout training, the teacher model is frozen and only the student model is updated via backpropagation.

%To suppress noisy supervision, we apply a cycle consistency check. Specifically, we track the final point $\hat{P}_i^T$ backward through the video to recover an estimate $\tilde{P}_i^1$ of the original location. The trajectory is retained only if the cycle consistency error is below a threshold $\alpha$:
% \begin{equation}
% \text{CCE}(i) = \left\| \tilde{P}_i^1 - q_i \right\|_2 < \alpha
% \end{equation}
% Trajectories satisfying this constraint are used as pseudo-labels to supervise the student. Only the student model is updated during training, while the teacher remains frozen.
%To reduce the impact of noisy supervision, we apply a cycle consistency check. Specifically, we start from the final point of each candidate trajectory and track it backward through the video. The recovered position in the first frame is then compared to the original query point. If the cycle consistency error exceeds a predefined threshold $\alpha$, the trajectory is considered unreliable and discarded. The remaining, temporally consistent trajectories are used as pseudo-labels to supervise the student. Throughout training, all teacher models are kept frozen, and only the student model is updated. \todo{Do you want to include a specific formula for the cycle consistency? Include some citation?}

\subsubsection{Student Fine-Tuning}
We train the student using supervision from both visible and occluded trajectories, following the loss formulation in CoTracker3 \cite{karaev2024cotracker3}. Tracking supervision is provided via a Huber loss with a threshold of 6, applied across multiple refinement iterations. To emphasize visible points, higher weight is assigned to their loss terms, while occluded points are down-weighted by a factor of 1/5. An exponential discount factor \(\gamma \in (0, 1)\) is also applied, reducing contribution of earlier iterations and encouraging accurate predictions in final refinement steps. The overall loss is defined as:

\begin{equation}
\mathcal{L}_{\text{track}} = \sum_{k=1}^{K} \gamma^{K - k} \left( \mathbf{1}_{\text{vis}} + \frac{1}{5} \mathbf{1}_{\text{occ}} \right) \cdot \text{Huber}(\mathcal{P}^{(k)}, \mathcal{P}^\star)
\end{equation}

where \( \mathcal{P}^{(k)} \) is the student’s prediction at refinement iteration \( k \), and \( \mathcal{P}^\star \) is the pseudo-label provided by teacher $M$. Since pseudo-labels can be noisy, we found it more stable to omit confidence and visibility supervision during fine-tuning. 
%Instead, we retain a fixed linear head (pretrained on synthetic data) to estimate confidence and visibility, which remains frozen throughout pseudo-label training. 
This helps prevent overfitting to unreliable label quality and focuses learning on trajectory refinement.
\section{Experiments}

\subsection{Datasets and Metrics}
We train on the Cholec80 dataset~\cite{twinanda2016endonet}, which contains 80 laparoscopic cholecystectomy videos exhibiting diverse anatomy, motion patterns, lighting conditions, and tool interactions. The videos are recorded at 25 FPS with an average duration of 2,306 seconds. Although it lacks point-level annotations, we use it as an unlabeled dataset for semi-supervised training. For evaluation, we use the STIR benchmark~\cite{zhan2024tracking}, comprising around 425 in-vivo and ex-vivo surgical videos recorded with a da Vinci Xi robot and annotated with over 3,000 points in the first and last frames of each sequence.
% We train on the Cholec80 dataset~\cite{twinanda2016endonet}, which comprises 80 laparoscopic cholecystectomy videos exhibiting diverse anatomy, motion patterns, lighting conditions, and tool interactions. The videos have an average length of 2,306.27 seconds, and are recorded at 25 FPS. Although it lacks point-level annotations, we use it as unlabeled data for pseudo-label generation in semi-supervised training. For evaluation, we use the STIR benchmark~\cite{zhan2024tracking}, which includes approximately 425 in-vivo and ex-vivo surgical videos recorded with a da Vinci Xi robot and annotated with over 3000 points in the first and last frames of each sequence. 
We filter out around 20 sequences with excessive label noise to ensure consistent evaluation.
% For training, we use the Cholec80 dataset \cite{twinanda2016endonet}, which contains 80 laparoscopic cholecystectomy videos capturing diverse surgical scenes with variations in anatomy, motion, lighting, and tool interaction. Although it lacks point-level annotations, we use it as unlabeled data to generate pseudo-labels for semi-supervised training. 
% For evaluation, we use the STIR dataset \cite{zhan2024tracking}, consisting of approximately 425 in vivo and ex vivo surgical videos with around 3000 labelled points recorded with a da Vinci Xi robot. Each sequence includes point annotations in the first and last frames. We filter out around 20 sequences with excessive label noise to ensure evaluation reliability. 

We evaluate tracking performance using three metrics: Mean Endpoint Error (MEE), Mean Chamfer Distance (MCD), and Average Accuracy $< \delta^{x}_{avg}$, as defined in TAP-Vid \cite{doersch2022tap}. The $< \delta^{x}_{avg}$ metric is computed as the average percentage of tracked points falling within thresholds of \{4, 8, 16, 32, 64\} pixels from ground truth positions.

\subsection{Implementation Details}

The student model is trained for 120{,}000 iterations using the Adam optimizer with a cosine learning rate schedule starting at $5 \times 10^{-5}$. Each batch contains a randomly sampled sequence with 64 query points tracked over 16 frames, sampled with a random stride between 1 and 4. Training is conducted on NVIDIA RTX 4090 GPUs. The cycle consistency  threshold $\alpha = 5$ provides the best trade-off between label quality and training signal. 
%We train the student model for 120{,}000 iterations using the Adam optimizer and a cosine learning rate schedule, with a base learning rate of $5 \times 10^{-5}$. We sample 64 points from sequences with 4 frames sampled from videos at a random interval ranging from 1-4.
%Training is conducted on NVIDIA RTX 4090 GPUs. To control the quality of pseudo-labels, we perform ablations over the cycle consistency threshold $\alpha$ and find that $\alpha = 5$ 

\subsection{Results}
We evaluate SurgTracker on the STIR benchmark, comparing it to several recent methods for point tracking, including RAFT \cite{teed2020raft}, SENDD \cite{schmidt2023sendd}, TAPIR \cite{doersch2023tapir}, BootsTAP ~\cite{doersch2024bootstap}, and CoTracker3 (Online)~\cite{karaev2024cotracker3}. As shown in Table ~\ref{table_results}, SurgTracker outperforms all baselines across all metrics. Compared to CoTracker3 (Online), which serves as our initialization and frozen teacher, it reduces MEE by 0.74 and MCD by 0.69 while improving $< \delta^{x}_{avg}$  by 0.44. These gains demonstrate the effectiveness of filtered self-distillation for adapting point trackers to high-shift surgical domains.

\begin{figure}[t]  
    \centering 
    \includegraphics[width=0.47\textwidth]{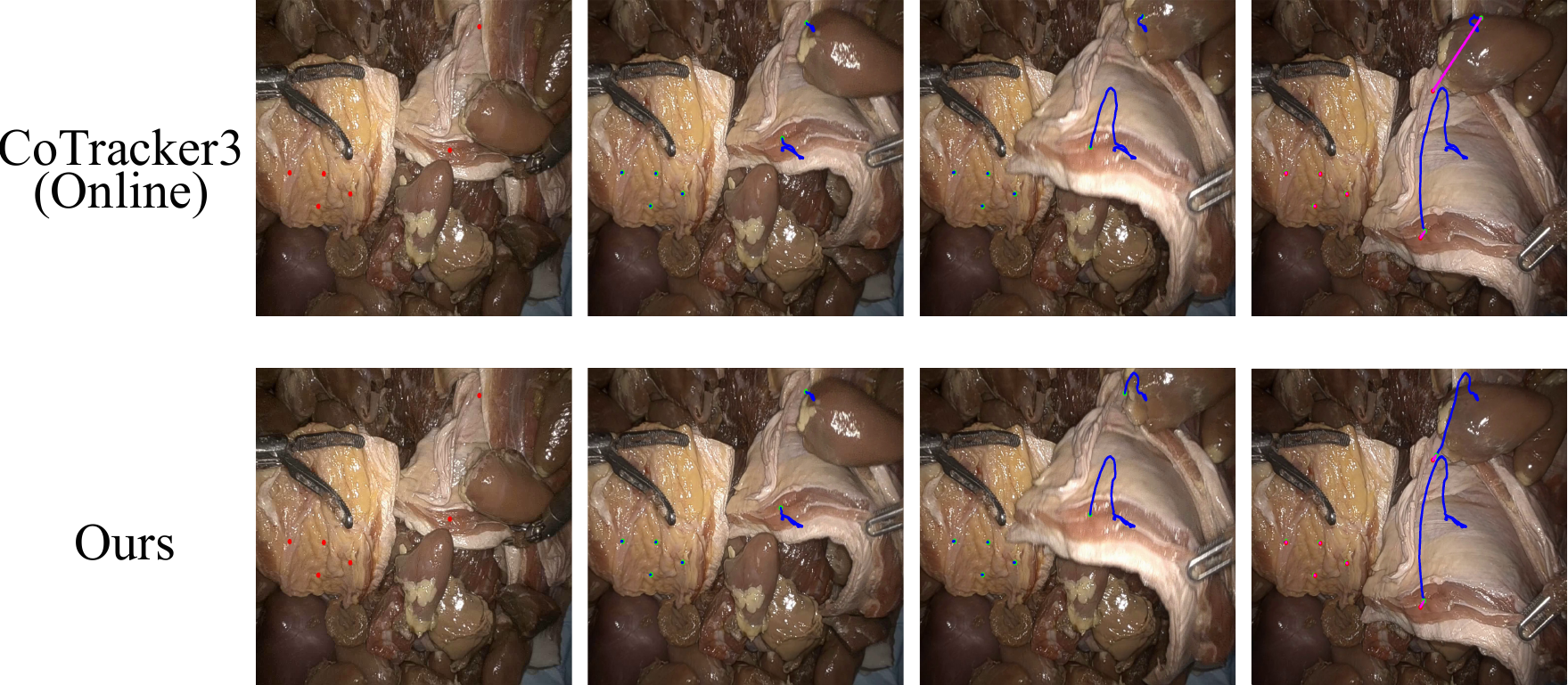}  
    \caption{Comparison of CoTracker3 and our method on a challenging sequence. Red and green dots mark initial and mid-frame predicted positions, respectively, blue lines show trajectories, and pink lines indicate final error. Our model better handles occlusion and motion change, accurately recovering the original trajectory.} 
    \label{fig:results} 
\end{figure}

Figure~\ref{fig:results} shows a qualitative comparison with CoTracker3 on a challenging occlusion scenario. Both models initially track the top-right point correctly until an occlusion occurs (second column). Notably, during the occlusion, the direction of the intended motion changes. While CoTracker3 drifts and continues tracking the occluding tissue with an estimation of the prior motion, our model successfully recovers and resumes tracking the original structure after it reappears (third column). The final trajectory is significantly more accurate, highlighting the robustness of our distilled model to occlusions and motion changes. 

\begin{table}[t]
\caption{Performance comparison on the STIR dataset.}
\centering
\begin{tabular}{c|c|c|c}
\hline
Method & MEE $\downarrow$ & MCD $\downarrow$ & $< \delta^{x}_{avg}$ $\uparrow$ \\
\hline
RAFT & 44.25 & 43.60 & 50.41\\
SENDD & 22.80 & 45.18 & 66.5 \\
TAPIR & 24.33 & 25.03 & 61.04\\
BootsTAP &20.38 & 21.4 & 63.74 \\
CoTracker3 (Online) & 17.01 & 17.81 & 68.11\\
\textbf{SurgTracker (Ours)} &\textbf{16.27} &\textbf{17.12} &\textbf{68.55} \\
\hline
\end{tabular}

\label{table_results}
\end{table}
\subsection{Ablation Studies}
To evaluate the impact of cycle consistency filtering, we vary the threshold $\alpha$ controlling the maximum allowed deviation between a point and its cycle-tracked counterpart. As shown in Table \ref{table_consistency}, omitting the filter results in lower accuracy, confirming the presence of noisy pseudo-labels. Filtering with $\alpha = 5$ achieves the best trade-off, minimizing both MEE and MCD while improving $< \delta^{x}_{avg}$ , reflecting more accurate tracking. A lower threshold ($\alpha$ = 2.5) is overly conservative, discarding too many training samples and thus limiting supervision. Conversely, a higher threshold ($\alpha$ = 7.5) allows more trajectories but admits additional noise, slightly degrading performance. These results highlight the importance of temporal consistency in improving label quality and overall tracking performance.

%Table \ref{table_teachers} compares different teacher configurations for pseudo-label generation. 
%Following CoTracker3, we evaluate multi-teacher setups combining CoTracker3 (Online), CoTracker3 (Offline), and BootsTAP in Table \ref{table_teachers}. In each configuration, a teacher model is randomly sampled from the pool of models in that setup in every batch and used to generate pseudo-labels for fine-tuning the student. 
Table~\ref{table_teachers} compares different teacher configurations for pseudo-label generation, following multi-teacher setup in CoTracker3. For each configuration, a teacher model is randomly sampled from corresponding pool per batch and used to generate pseudo-labels for student fine-tuning.
Our self-distillation approach, which uses only CoTracker3 (Online) as the teacher, achieves the best performance. Incorporating other teachers slightly degrades performance, likely due to inconsistent supervision that hinders stable learning. These findings suggest that, under a significant domain shift, a consistent, architecture-aligned teacher can outperform diverse ensembles, offering more effective supervision.
\subsection{Conclusion}
We present SurgTracker, a semi-supervised framework for adapting synthetic-trained point trackers to surgical video through filtered self-distillation. By leveraging a single, architecture-aligned teacher and enforcing cycle consistency, our method provides stable, high-quality supervision without the overhead of maintaining teacher ensembles. Experiments on the STIR benchmark demonstrate that SurgTracker improves tracking performance using only 80 unlabeled videos, demonstrating that consistent supervision can outperform diverse teacher setups in challenging, high-shift surgical domains.
%We propose SurgTracker, a semi-supervised framework that adapts synthetic-trained point trackers to surgical video using filtered self-distillation. By leveraging a single, architecture-aligned teacher with cycle consistency filtering, our method improves tracking performance while reducing memory overhead compared to multi-teacher setups. This highlights the importance of consistent supervision in high-shift and data-scarce domains.

\begin{table}[t]
\caption{Ablation on cycle consistency threshold $\alpha$.}
\centering
\begin{tabular}{c|c|c|c}
\hline
$\alpha$ & MEE $\downarrow$ & MCD $\downarrow$ & $< \delta^{x}_{avg}$ $\uparrow$ \\
\hline
No filtering &16.69 &17.46 &68.04 \\

2.5 & 16.76 & 17.58 & 68.02 \\

5 & \textbf{16.27} &\textbf{17.12} &\textbf{68.55} \\

7.5 & 16.43 & 17.23 & 68.31\\
\hline
\end{tabular}
\label{table_consistency}
\end{table}

\begin{table}[t]
\caption{Comparison of different teacher configurations. The student is always CoTracker3 (Online). CoT3 (On) and CoT3 (Off) refer to online and offline versions of CoTracker3 respectively.}
\centering
\begin{tabular}{c|c|c|c}
\hline
Teacher Models & MEE $\downarrow$ & MCD $\downarrow$ & $< \delta^{x}_{avg}$  $\uparrow$ \\
\hline
CoT3 (On) & \textbf{16.27} &\textbf{17.12} &\textbf{68.55} \\
\hline
\begin{tabular}[c]{@{}c@{}}CoT3 (On), CoT3 (Off)\end{tabular} &16.28 &17.10 &68.50 \\
\hline
\begin{tabular}[c]{@{}c@{}}CoT3 (On), CoT3 (Off), \\ BootsTAP \end{tabular} & 16.38 & 17.21 & 68.39 \\
\hline
\begin{tabular}[c]{@{}c@{}}CoT3 (On), CoT3 (Off), \\ Track-On \end{tabular} & 16.80 & 17.64 & 68.00 \\
\hline
\end{tabular}
\label{table_teachers}
\end{table}
\section{Acknowledgments}
The work described in this paper was conducted in the framework of the Graduate School 2543/1 “Intraoperative Multi-Sensory Tissue Differentiation in Oncology" (project ID 40947457) funded by the German Research Foundation (DFG - Deutsche Forschungsgemeinschaft). This work has been supported by the Deutsche Forschungsgemeinschaft (DFG) – EXC number 2064/1 – Project number 390727645. The authors thank the International Max Planck Research School for Intelligent Systems (IMPRS-IS) for supporting Valay Bundele and Mehran Hosseinzadeh.

{
    \small
    \bibliographystyle{ieeenat_fullname}
    \bibliography{main}
}


\end{document}